\documentclass[10pt,twocolumn,letterpaper]{article}

\usepackage{iccv}
\usepackage{times}
\usepackage{epsfig}
\usepackage{graphicx}
\usepackage{amsmath}
\usepackage{amssymb}

\usepackage{float}
\usepackage{booktabs}
\usepackage{soul}
\usepackage[accsupp]{axessibility}

\usepackage[pagebackref=true,breaklinks=true,letterpaper=true,colorlinks,bookmarks=false]{hyperref}

\usepackage[capitalize]{cleveref}
\crefname{section}{Sec.}{Secs.}
\Crefname{section}{Section}{Sections}
\Crefname{table}{Table}{Tables}
\crefname{table}{Tab.}{Tabs.}

\iccvfinalcopy % *** Uncomment this line for the final submission

\def\iccvPaperID{****} % *** Enter the ICCV Paper ID here

\ificcvfinal\fi

\begin{document}

%%%%%%%%% TITLE
\title{Shape Anchor Guided Holistic Indoor Scene Understanding}

% \texttt{shshen@nlpr.ia.ac.cn}}

\author{Mingyue Dong$^{1}$, ~ Linxi Huan$^{1}$, ~Hanjiang Xiong$^{1}$,  ~Shuhan Shen$^{2}$, ~Xianwei Zheng$^{1}$\thanks{Corresponding author.}\\
$^1$The State Key Lab. LIESMARS, Wuhan University\\ $^2$Institute of Automation, Chinese Academy of Sciences \\
\texttt{\{dmy25148, whu\_hlx, xionghanjiang, zhengxw\}@whu.edu.cn} \\
\texttt{shshen@nlpr.ia.ac.cn}}

\maketitle
% Remove page # from the first page of camera-ready.
\ificcvfinal\thispagestyle{empty}\fi

%%%%%%%%% ABSTRACT
\begin{abstract}
This paper proposes a shape anchor guided learning strategy (AncLearn) for robust holistic indoor scene understanding. We observe that the search space constructed by current methods for proposal feature grouping and instance point sampling often introduces massive noise to instance detection and mesh reconstruction. Accordingly, we develop AncLearn to generate anchors that dynamically fit instance surfaces to (i) unmix noise and target-related features for offering reliable proposals at the detection stage, and (ii) reduce outliers in object point sampling for directly providing well-structured geometry priors without segmentation during reconstruction. We embed AncLearn into a reconstruction-from-detection learning system (AncRec) to generate high-quality semantic scene models in a purely instance-oriented manner. Experiments conducted on the challenging ScanNetv2 dataset demonstrate that our shape anchor-based method consistently achieves state-of-the-art performance in terms of 3D object detection, layout estimation, and shape reconstruction. The code will be available at \href {https://github.com/Geo-Tell/AncRec}{https://github.com/Geo-Tell/AncRec}.
\end{abstract}

\vspace{-1em}
\section{Introduction}
\label{sec:intro}

Holistic indoor scene understanding from partial observations (\eg, single-view images or 3D scans) is a comprehensive task that provides 3D semantic scene models for indoor applications. Early works studied this task with a reconstruction-from-detection framework that recovers the geometries of room structures and objects from the corresponding detection in a separate way. Later, end-to-end learning methods were proposed to simultaneously perform layout estimation, object detection, and shape prediction in one forward pass for semantic scene reconstruction \cite{nie2020total3dunderstanding,zhang2021holistic,huan2022georec}. With the recent success of point-based 3D detection and instance reconstruction, surging interest has been witnessed in detecting and modeling objects directly from sparse point clouds \cite{nie2021rfd,tang2022point}. 

\begin{figure}[t]
 \setlength{\belowcaptionskip}{-2em}
    \centering
    \includegraphics[width=\linewidth]{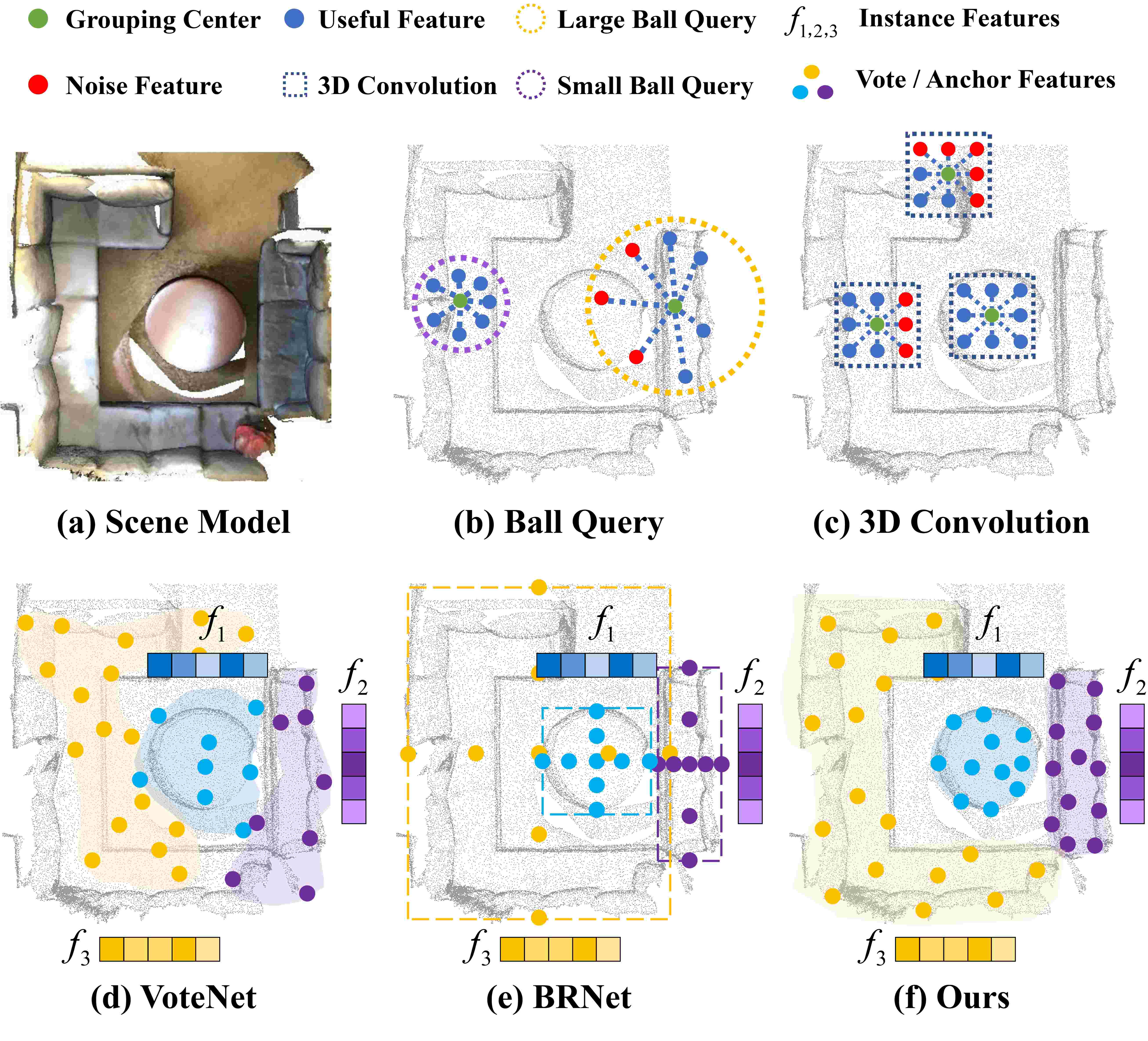}
    \caption{Comparison between different feature grouping strategies. (a) the original scene model. (b)-(e) The different feature grouping operations all suffer from the issue of confusing non-target noise with useful features. (f) The proposed shape anchor guided grouper directly generates anchors at the instance surface to merge instance-related features, which largely alleviates noise interference.}
    \label{fig:ins_info_learn}
\end{figure}

 Benefiting from the use of rich geometry information, current scan-based deep methods have improved the performance of semantic scene reconstruction. However, they still left two issues that bottleneck high-quality semantic reconstruction: (1) the noisy instance feature learning at the detection phase, and (2) the difficulty in retrieving instances from sparse point clouds for reconstruction. 
 
 At the detection phase, it is required to group features for instance representation learning. The ball query \cite{qi2017pointnet,qi2017pointnet++} and 3D convolution\cite{zhou2018voxelnet,hou20193d} are two basic operations for point feature grouping, but they often mix massive noise with informative features due to the fixed grouping range, as shown in \cref{fig:ins_info_learn} (b) and (c). To compensate for the deficiency caused by the fixed range, VoteNet  \cite{ding2019votenet} and its variants \cite{xie2021venet, qi2020imvotenet, zhang2020h3dnet, xie2020mlcvnet} adopt a voting strategy to cluster object  features by moving surface points towards object centers. Albeit more flexible than the basic operations, the voting-based strategy often generates an unconstrained grouping area that brings quantities of outliers as illustrated in \cref{fig:ins_info_learn} (d). BRNet \cite{cheng2021back} hence restricts the grouping space by sampling around representative points  given by coarse box proposals. Nevertheless, limited by box-like grouping area, the sampling points can still fall far beyond targets when the objects are irregularly shaped, as depicted in \cref{fig:ins_info_learn} (e).

 At the reconstruction stage, the retrieval of outlier-free object points is a prerequisite for object recovery. Due to the noise introduced by feature grouping in the previous detection phase, the points grouped for localizing objects can hardly serve as ideal reconstruction priors as indicated by \cref{fig:ins_info_learn}. Consequently, the current methods are forced to employ an additional foreground classifier \cite{nie2021rfd} or replace the detector with a complex instance segmentation backbone to sample object points from the raw scans \cite{tang2022point}. However, the existence of numerous non-target outliers in the search space challenges instance segmentation, resulting in increased risks of gluing different instances and misclassifying background points.

 Based on the discussion above, the two noise interference issues during detection and reconstruction are actually highly coupled and can be resolved together as long as outliers are excluded during feature grouping. To this end, we are motivated to propose a shape anchor-guided learning strategy (\emph{AncLearn}) that generates surface anchors to fit the feature grouping areas to object shape distributions, as displayed in \cref{fig:ins_info_learn} (f). With the geometry constraint provided by surface anchors, it is able to merge local target-focused features for predicting reliable object proposals and construct a shape-aware search space for robustly sampling instance points without segmentation during reconstruction. The proposed anchor-guided learning strategy can be easily embedded into an end-to-end learning system to accomplish object detection, layout estimation, and instance reconstruction for holistic scene understanding.
 The main contributions are summarized as follows:
 % \vspace{-0.5em}
\begin{enumerate}
    \item[-] We present a shape anchor guided learning strategy to simultaneously address the issues of noisy feature learning in detection and instance point retrieval during reconstruction.
    
    %learn informative instance features with less noise interference for scene parsing, and provide instructive geometry priors for object reconstruction.
 \vspace{-0.5em}
    
    \item[-] We embed the proposed anchor-guided learning strategy into an end-to-end learning system to accomplish object detection, layout estimation, and instance reconstruction for holistic scene understanding in a purely instance-oriented way.
 \vspace{-0.5em}
    
    \item[-] Extensive experiments demonstrate that our AncRec achieves high-quality semantic scene reconstruction with state-of-the-art performance in instance detection and mesh prediction on the challenging ScanNetv2 dataset \cite{dai2017scannet} (some ground truths provided by Scan2CAD \cite{avetisyan2019scan2cad} and SceneCAD \cite{avetisyan2020scenecad}).
 \end{enumerate}

\begin{figure*}[t]
\setlength{\belowcaptionskip}{-0.5em}
    \centering
    \includegraphics[width=\linewidth]{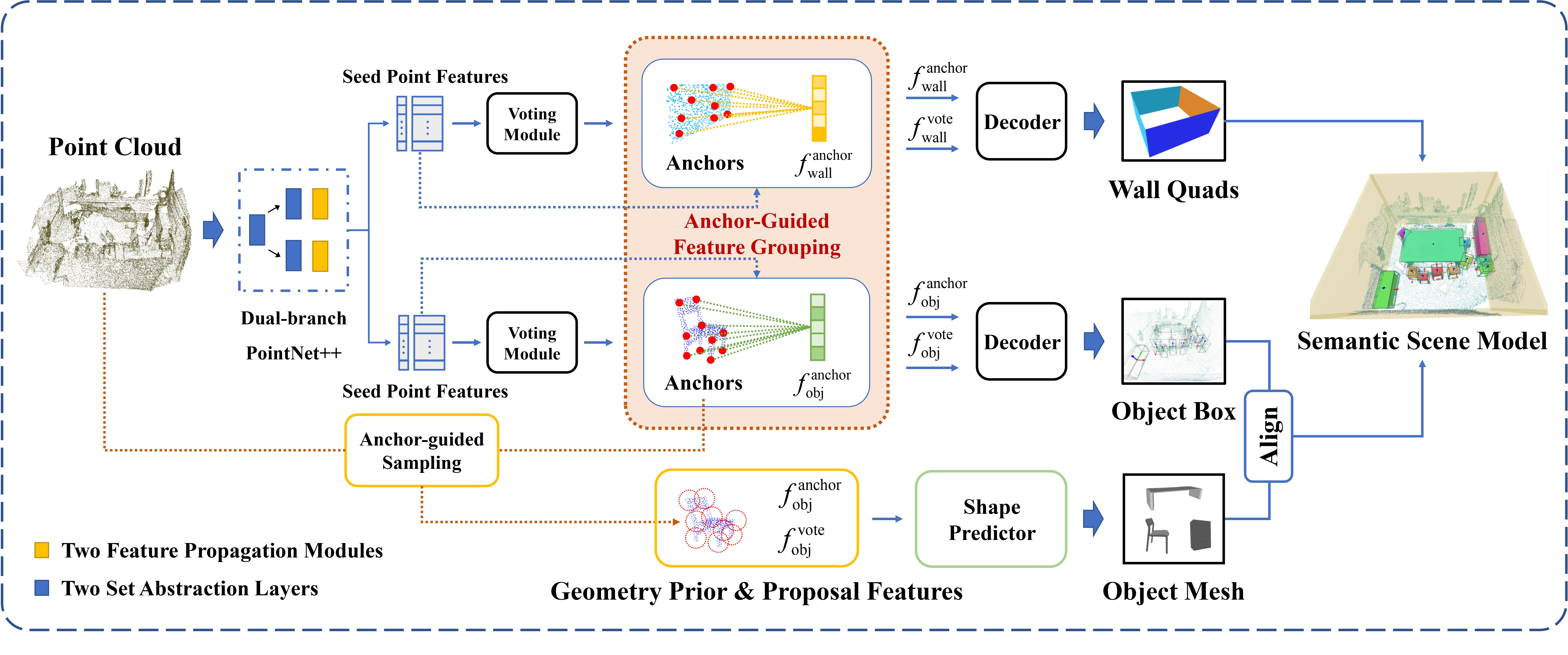}
    \caption{Overview of the AncRec framework. At the detection stage, seed point features for walls and objects are first learned with a modified dual-branch PointNet++. Following each branch, a voting module and an anchor-guided feature grouper respectively generate proposal features $\boldsymbol{f}_\text{obj(wall)}^\text{vote}$ and $\boldsymbol{f}_\text{obj(wall)}^{\text{anchor}}$, which are fed into the decoders to predict object bounding boxes and wall quads. The room layout is then constructed by processing the wall quads into connected corners. At the reconstruction stage, objects with high objectness scores are reconstructed under the guidance of $\boldsymbol{f}_\text{obj}^{\text{vote}}$, $\boldsymbol{f}_\text{obj}^{\text{anchor}}$, and the geometry priors sampled by shape anchors. Finally, object models are arranged in the scene with the predicted layout according to the spatial alignment provided  by predicted bounding boxes.}
    \label{fig:anchorrec}
    \vspace{-1em}
\end{figure*}

\vspace{-0.5em}
\section{Related Work}
\label{part:related-work}
\vspace{-0.5em}
Semantic scene understanding has been extensively studied over the past years. Many methods focused on acquiring the semantics of 3D scenes \cite{hou2021exploring, engelmann20203d, chen20224dcontrast},  while others recovered scene geometries by scene completion \cite{han2019deep,dai2020sg,yin2022towards}. Recently,  increasing interest has emerged in semantic scene reconstruction, which recovers both the semantics and geometric shapes of objects. By treating semantic reconstruction as a problem of holistic scene understanding, promising progress has been made based on a reconstruction-from-detection principle \cite{nie2021rfd}. In the following, we review the research from the two core aspects of the reconstruction-from-detection pipeline, \ie, 3D object detection and scene-aligned instance reconstruction.

\subsection{3D Object Detection}
Over the past years, deep detectors have gained great success in 2D object detection \cite{zou2023object,liu2020deep}. The progress of 2D deep detection has inspired the development of deep learning techniques for recognizing objects from scene point clouds. Compared to image data,  point clouds provide rich surface geometry clues for locating objects in real scenes. Nevertheless, the sparse, irregular, and orderless characteristics of point clouds make them hard to be handled by the grid-based convolution model. 

Works in the early stage leveraged the 2D proposals as 3D detection constraints or projected point clouds into regular 2D/3D grids. Although these 3D detectors are  applicable, their performance is still limited by the  2D detectors and the missing geometric details brought by projection. To directly learn the rich geometry features from the raw points, PointNet \cite{qi2017pointnet} and PointNet++ \cite{qi2017pointnet++} used a ball query operation for grouping point features. Later, PointRCNN \cite{shi2019pointrcnn} adopted the point-wise features extracted by PointNet++ for producing instance proposals with point clouds. Hindered by the fixed feature grouping range of grid-based convolutions or ball query operation, these methods often mixed massive noise with informative features, which impairs the reliability of proposals. Considering that the observed object surface points usually lie far away from the object centers, VoteNet \cite{ding2019votenet} introduced a voting mechanism for proposal generation. Based on the voting mechanism, numerous variants further refined proposal features with context relationships \cite{xie2021venet}, hierarchical clues \cite{xie2020mlcvnet}, and hybrid geometric primitives \cite{zhang2020h3dnet}. However, the issue of outliers in votes still blocks the learning of representative box features. BRNet \cite{cheng2021back} derived proposal features based on the generated virtual points given by coarse box proposals, whereas the virtual points can fall into non-target areas and bring noise. In this paper, we directly merge features anchored on target surfaces into robust proposal representations for indoor instance learning.

\subsection{Scene-aligned Instance Reconstruction}
Scene-aligned instance reconstruction requires not only modeling 3D object shapes but also correctly arranging the shapes in the 3D scene space. The semantic model of an indoor scene was early constructed with retrieval techniques, which search for CAD shapes or geometric primitives from an offline database and align the approximate object models to input scene data \cite{avetisyan2019scan2cad,avetisyan2019end,gupta2015aligning,izadinia2017im2cad}. % Automatic semantic modeling of indoor scenes from low-quality RGB-D data using contextual information; A search-classify approach for cluttered indoor scene understanding; Acquiring 3D indoor environments with variability and repetition; An Interactive Approach to Semantic Modeling of Indoor Scenes with an RGBD Camera; Complete 3D Scene Parsing from an RGBD Image
% The retrieval process was later investigated as a problem of deep feature matching with convolutional neural networks (CNNs)  \cite{}. 
Albeit able to present delicate scene models, the retrieval-based methods lack generalization ability to various scenes due to the inefficient inference and the limited database scale \cite{zhang2021holistic}.

The promising advances in deep learnable shape representations motivated scene-aligned instance reconstruction without a finite model pooling \cite{genova2020local,chibane2020implicit}.	Many prior arts in literature generate 3D objects with explicit or implicit representations with learned features derived from 2D recognition results \cite{hanocka2019meshcnn,nie2020total3dunderstanding,zhang2021holistic,huan2022georec}. Unlike previous single-view modeling methods, RevealNet  \cite{Hou2019RevealNetSB} and the following works  \cite{bokhovkin2021towards, bokhovkin2023neural}  predicted the occupancy grids of semantic instances with 3D features extracted from voxelized RGB-D scans. For high scene resolution that is hard to acquire by volumetric representation, RfD-Net \cite{nie2021rfd} and DIMR \cite{tang2022point} extracted instance points from scans with segmentation techniques for subsequent single object reconstruction with deep implicit functions \cite{mescheder2019occupancy,chen2020bsp}.  Because of the existence of non-target points, these two state-of-the-art methods have to use an extra foreground classifier or a complex instance segmentation backbone to sample object points for reconstruction. In this paper, we utilized the previously generated surface anchors to localize instance points in scans. Hereby, we introduced an easy but effective anchor-guided sampling strategy to offer geometry priors for the following instance reconstruction without an extra segmentation stage.

\vspace{-0.5em}
\section{Method}
\label{part:methodology}
\vspace{-0.5em}
We illustrate the proposed AncRec in \cref{fig:anchorrec}.  AncRec achieves semantic scene reconstruction with the proposed anchor-guided learning strategy in an instance-oriented way. It first localizes objects and walls via a dual-branch detector that is equipped with shape anchors learned for grouping point features at the detection stage. The object shape anchors are subsequently leveraged to sample instance geometry priors for predicting object meshes. Finally, the complete semantic scene model is reconstructed by arranging object meshes in the post-processed room layout with alignment to the parsed bounding boxes and poses. In the following, we elaborate on how the shape anchors are learned and leveraged to address the noise interference during instance detection and mesh prediction for high-quality semantic scene reconstruction.

\subsection{Shape Anchor-guided Instance Detection}
\label{part:detector}
We modify the VoteNet into a dual-branch instance detector to simultaneously localize objects and walls in one forward pass (\cref{fig:anchorrec}). Although the voting mechanism of VoteNet is workable for generating 
 proposals, the reliability of voting-based predictions is often impaired by the outliers in votes. We hence design an anchor-guided feature grouper to learn target-focused features for robust instance detection.

\vspace{-1em}
\subsubsection{Anchor-guided Feature Grouper}
\label{sec:Anchor-guided Feature Grouper}
\begin{figure}[h]
\setlength{\belowcaptionskip}{-1.5em}
    \centering
    \includegraphics[width=\linewidth]{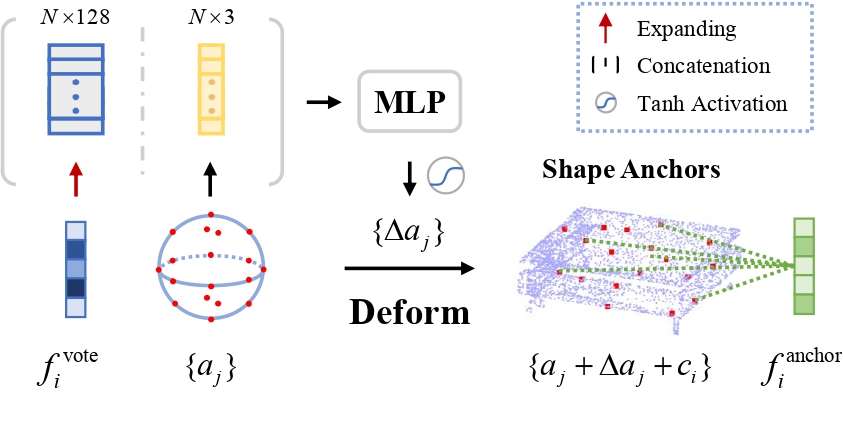}
    \caption{The anchor-guided feature grouper. The anchor-guided proposal feature $\boldsymbol{f}_{i}^{\text{anchor}}$ is the average of features at shape anchors. The shape anchors are deformed from initial anchors $\left\{\boldsymbol{a}_j\right\}$ with offsets derived from  the concatenation of the voting-based proposal feature $\boldsymbol{f}_{i}^{\text{vote}}$ and $\left\{\boldsymbol{a}_j\right\}$. The anchors are then translated to the cluster center $\boldsymbol{c}_i$.}
    \label{fig:anchor_grouper}
\end{figure}

\vspace{-0.5em}
We illustrate the mechanism of the anchor-guided feature grouper in \cref{fig:anchor_grouper}. In the following, by taking the object detection branch as an example, we describe how it works with the voting module in three steps to predict the oriented 3D bounding boxes. 

\textbf{Step 1: Anchor Generation.} Given the $i^{th}$  voting-based proposal feature $\boldsymbol{f}^{\text{vote}}_{i}\in\mathbb{R}^{128}$, the anchor-guided grouper first generates shape anchors that depict the geometry of the $i^{th}$ object candidate via template deformation. With $N$ initial anchors $A=\left\{\boldsymbol{a}_j \in \mathbb{R}^3| j=1,...,N\right\}$ uniformly selected from a unit ball surface, a deformation layer is applied to translate the initial anchors to the $i^{th}$ candidate object surface by
\begin{equation}
    \label{eq:anchor_gen}
    \hat{\boldsymbol{a}}_j = \boldsymbol{a}_j + \underbrace{\text{Tanh}(\text{MLP}([\boldsymbol{a}_j, \boldsymbol{f}^{\text{vote}}_{i}]))}_{\text{deformation offset}\ \Delta \boldsymbol{a}_j} + \boldsymbol{c}_{i}.
\end{equation}

As indicated by \cref{eq:anchor_gen}, the deformation layer moves $\boldsymbol{a}_j$ with offsets $\Delta \boldsymbol{a}_j$ inferred by a multi-layer perceptron (MLP) and a Tanh activation function and then places the deformed anchors at the candidate center $\boldsymbol{c}_i$ clustered by the voting module. The training of the deformation layer is supervised by the Chamfer distance loss defined as 
\begin{equation}
\setlength{\abovedisplayskip}{0.2em}
    \label{eq:chamfer}
    \begin{aligned}
        \mathcal{L}_{\mathrm{anchor}}=&\frac{1}{|P_{\text{gt}}|} \sum_{\boldsymbol{p} \in P_{\text{gt}}} \min _{\boldsymbol{\hat{a}} \in \hat{A}}\|\boldsymbol{p}-\boldsymbol{\hat{a}}\|_{2}^{2}\\
        &+\frac{1}{|\hat{A}|} \sum_{\boldsymbol{\hat{a}} \in \hat{A}} \min _{\boldsymbol{p} \in P_{\text{gt}}}\|\boldsymbol{\hat{a}}-\boldsymbol{p}\|_{2}^{2},  		
    \end{aligned}
\end{equation}
where $\hat{A}$ and $P_{\text{gt}}$ denote the sets of deformed anchors and surface points sampled from the corresponding scene-aligned 3D meshes. Thereby, the anchors are learned to fit the $i^{th}$ object surface, and the clustering center $\boldsymbol{c}_i$ is also refined with shape constraint provided by \cref{eq:chamfer}. Moreover, due to the supervision with complete surface points, the shape anchors can assist in exploiting context for recovering unobserved structures, \eg, the missing object bottom that is supported by the observed floor areas.

\textbf{Step 2: Feature Grouping.} The anchor-guided grouper propagates the seed point features (extracted by the dual-branch PointNet++ backbone) to each anchor via interpolation followed by an MLP. As the deformed anchors are mainly located on the object surface, the seed point features for propagation can be reliably selected from target-related areas. The  $i^{th}$  noise-reduced proposal representation $\boldsymbol{f}^{\text{anchor}}_{i}\in\mathbb{R}^{128}$ is obtained by averaging the anchor features.  

\textbf{Step 3: Prediction Fusion.}
We employ two decoders to predict the object category and oriented box parameters from $\boldsymbol{f}^{\text{vote}}_{i}$ and $\boldsymbol{f}^{\text{anchor}}_{i}$ respectively  for the  $i^{th}$ candidate. This is based on the consideration that the voting-based proposal feature $\boldsymbol{f}^{\text{vote}}_{i}$ contains more contextual information while the anchor-guided proposal feature $\boldsymbol{f}^{\text{anchor}}_{i}$ is more target-focused. The estimated parameters, denoted as $\boldsymbol{\Theta}^{\text{vote}}_i$ and $\boldsymbol{\Theta}^{\text{anchor}}_i$, are averaged with learnable weights to obtain the final object parameters:
\begin{equation}
    \label{eq:prediction_fusion}
    \boldsymbol{\Theta_i} = \boldsymbol{w_1} \cdot \boldsymbol{\Theta}^{\text{vote}}_i + \boldsymbol{w_2} \cdot \boldsymbol{\Theta}^{\text{anchor}}_i.
\end{equation}

From the three steps above, the attributes of objects are parsed with robustness to noise. The anchor-guided detection of wall instances works in a similar way. Considering that walls are generally connected to each other, we additionally deploy the attention operation used in  \cite{xie2020mlcvnet} to enhance the anchor-guided wall proposal features with the strong relationship between walls for precise layout estimation.

\vspace{-1em}
\subsubsection{The Detection Training Loss}
\vspace{-0.5em}
The total loss function of the anchor-guided instance detector is defined as 
\begin{equation}
    \label{eq:detection_loss}
    \mathcal{L}=\mathcal{L}_{\text{obj}}+\mathcal{L}_{\text{wall}}
    +\mathcal{L}_{\text{obj}}^{\text{anchor}}
    +\mathcal{L}_{\text{wall}}^{\text{anchor}}.
\end{equation}
In \cref{eq:detection_loss}, $\mathcal{L}_{\text{obj}}$ is the object detection loss given by \cite{ding2019votenet} while $\mathcal{L}_{\text{wall}}$ is the wall quad loss used in \cite{chen2022pq}. $\mathcal{L}_{\text{obj}}^{\text{anchor}}$ and $\mathcal{L}_{\text{wall}}^{\text{anchor}}$ are the chamfer distance losses that supervise the anchor deformation learning in terms of positive object and wall candidates.

\subsection{Anchor-guided Object Reconstruction}
\label{part:reconstruction}
\begin{figure}[h]
 % \vspace{-1em}
    \centering
    \includegraphics[width=\linewidth]{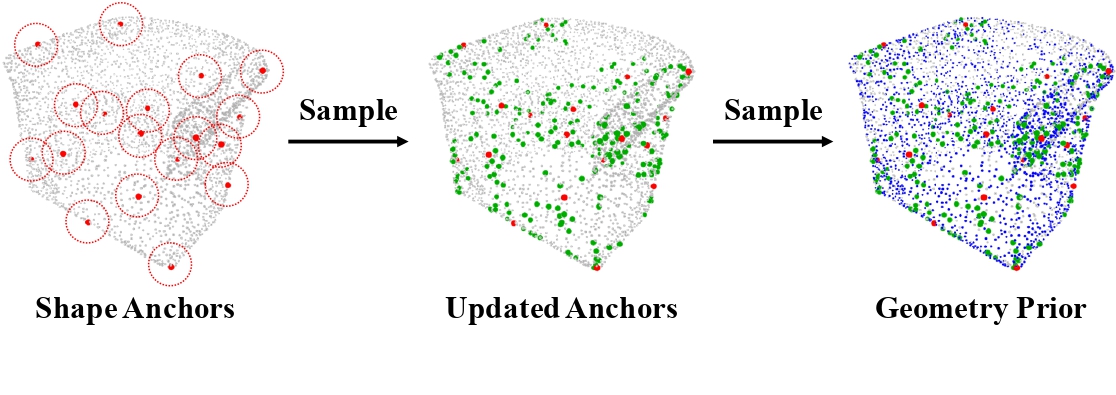}
    \vspace{-2em}
    \caption{The anchor-guided instance point sampling. Object points are iteratively sampled from the input scan under the guidance of shape anchors. The sampled object points serve as the geometry prior for shape reconstruction.}
    \label{fig:anchor_sampler}
\end{figure}

Instance points from the input scan are desirable geometry priors for reconstruction. Previous works utilized segmentation operations for instance point sampling. Despite being applicable, accurate instance segmentation is difficult due to the existence of massive non-target noise in the search space. In contrast, we take advantage of the shape anchors to construct a shape-aware search space, in which object points can be directly localized with little noise interference. \cref{fig:anchor_sampler} illustrates the workflow of our anchor-guided instance point sampling.

We first add the shape anchors of positive proposals to the original scene scan to enhance indoor objects' structural information. From the enhanced scan, we select points that lie within a given radius of each anchor and update the anchor set with these selected points to expand the searching space. With the searching space compactly fitting the object shape, we can further sample more instance points with high coverage. In this way, our anchor-guided sampling efficiently generates object geometry prior without reliance on an extra segmentation module. In the experiments, the anchor-guided sampling process iterates twice with the sampling radius set to the minimum distance between the anchors.

 % \vspace{-1em}
 
The geometry priors are next concatenated with the corresponding proposal features $\boldsymbol{f}^{\text{vote}}_{\text{obj}}$ and $\boldsymbol{f}^{\text{anchor}}_{\text{obj}}$  and encoded into shape embeddings  $\boldsymbol{f}^{\text{shape}}_{\text{obj}}$ by a ResPointNet \cite{nie2021rfd,qi2017pointnet}.   Based on $\boldsymbol{f}^{\text{shape}}_{\text{obj}}$, the decoder of BSP-Net \cite{chen2020bsp} is adopted to predict the signed distances of query points at the canonical coordinates. Following the BSP-Net, we apply the Constructive Solid Geometry (CSG) technique to extract the shape surfaces.

\vspace{-0.2em}
\subsection{Semantic Scene Reconstruction}
\vspace{-0.5em}
We now build the semantic scene model with the anchor-guided results of the instance detector and shape predictor. The room structure is first constructed by transforming the detected wall quads into orderly connected corners with the merging technique \cite{chen2022pq}. Next, the models of objects with high objectness are chosen to be arranged in the scene with alignment to the predicted 3D bounding boxes \cite{nie2021rfd}. In the end, the semantic scene model is built as a combination of the reconstructed layout and the scene-aligned object shapes.

\vspace{-0.5em}
\section{Experiments}
\label{sec:experiments}
% \vspace{-0.5em}
\subsection{Experiment Settings}
% \vspace{-0.5em}
\subsubsection{Dataset}
% \vspace{-0.5em}
We test our method on ScanNetv2 \cite{dai2017scannet} with ground truths from Scan2CAD \cite{avetisyan2019scan2cad} and SceneCAD datasets \cite{avetisyan2020scenecad} for holistic indoor scene understanding. (1) ScanNetv2 is a benchmark for indoor scene analysis with 1,513 scanned room point clouds; (2) Scan2CAD is an alignment dataset that matches ShapeNet models to their counterpart object instances in ScanNet with oriented 3D bounding boxes; and (3) SceneCAD provides 3D layout annotations for the scans in ScanNetv2. We follow \cite{chen2022pq} and \cite{chen2020bsp} to pre-process the layout polygons and object meshes for network training supervision. The train/test split is kept in line with the previous works with eight challenging object categories considered in the experiments.
\vspace{-0.5em}
\subsubsection{Evaluation Metrics}
\vspace{-0.5em}
The quality of semantic scene reconstruction is evaluated with the performance of object detection, layout estimation, and scene-aligned shape reconstruction. In line with previous works \cite{nie2021rfd,chen2022pq}, we use mean average precision across all classes (mAP) with 0.5 IoU threshold for object detection and F1-score for layout estimation. To evaluate the reconstruction quality, we use the chamfer distance (CD) based mAP with thresholds 0.1 and 0.047, light field distance (LFD) based mAP with thresholds 5000 and 2500, as well as the mAP at 3D IoU thresholds 0.25 and 0.5.
\vspace{-1em}
\subsubsection{Implementation Details}
\vspace{-0.5em}
The training of our method is conducted on a Titan GPU with two stages, and all parameters are updated by the Adam optimizer. In the first stage, we train the dual-branch instance detector of AncRec for simultaneous optimization of object detection and layout estimation modules. We set the  batch size to 8, initialize the learning rate to 1e-3, and adopt the \emph{ReducerLrOnPlateu} learning scheduler in PyTorch. In the second stage, we use the object proposals predicted by the instance detector for mesh reconstruction. We train the shape predictor with the BSPNet decoder pre-trained on the 8-category ShapeNet data \cite{chang2015shapenet} for training stability and efficiency. The training process lasts 100 epochs with batch size set to 32, learning rate initialized as 1e-4, and the same training schedule. We found the number of shape anchors $N$ is an insensitive hyper-parameter and thus set it to 18 for computational efficiency.

\subsection{Comparison and Analysis}
% \vspace{-0.5em}
\begin{figure*}[ht]
\centering
\begin{tabular}{lp{17cm}}
\multicolumn{2}{c}{\includegraphics[width=0.88\linewidth]{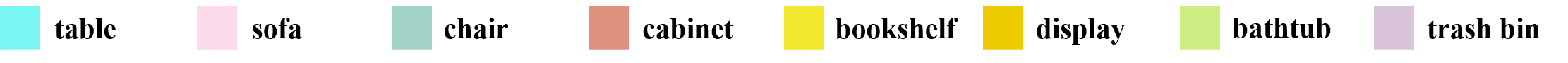}}\vspace{-0.9em}%\hspace{-9em}

\\
\rotatebox{90}{\hspace{-0.5cm}\small \textbf{GT}}&\hspace{-0.2cm}
    \begin{minipage}{\linewidth}
        \centering
        \includegraphics[width = 0.194\linewidth]{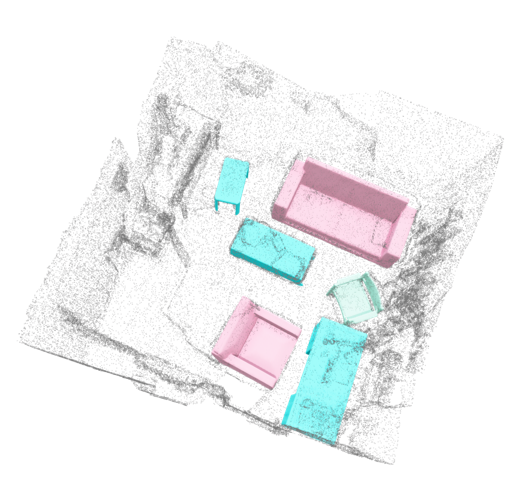}
        \includegraphics[width = 0.194\linewidth]{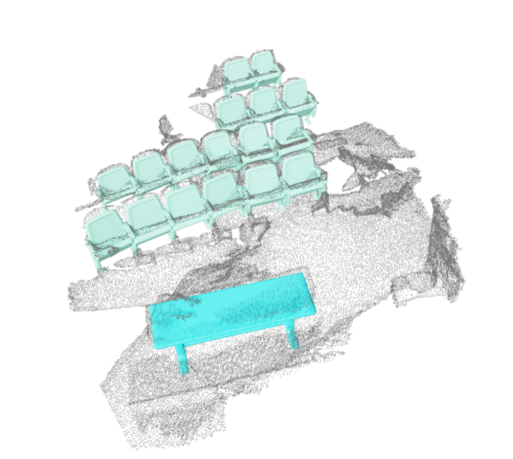}
        \includegraphics[width = 0.194\linewidth]{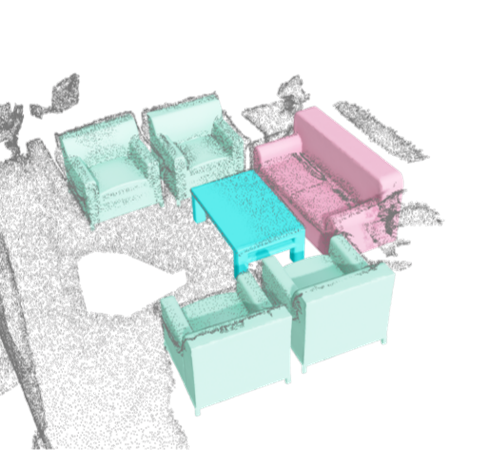}
        \includegraphics[width = 0.194\linewidth]{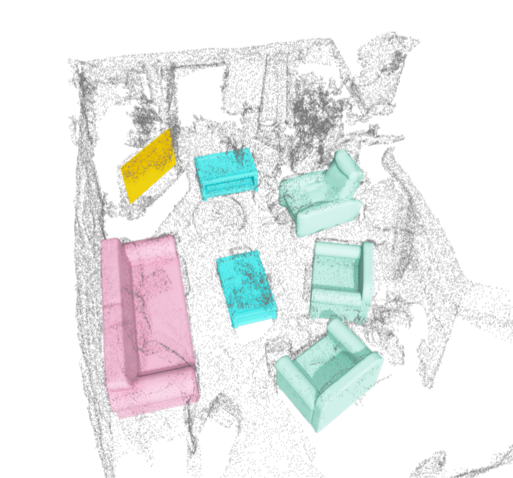}
        \includegraphics[width = 0.194\linewidth]{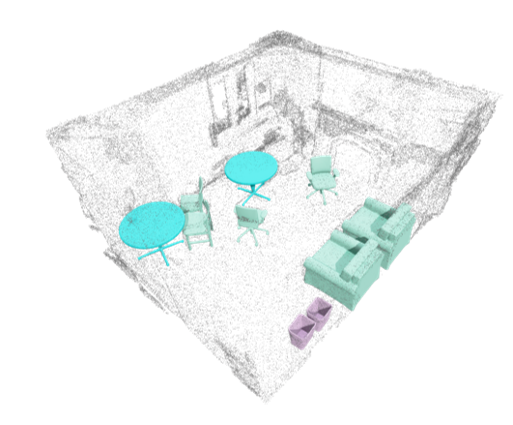}
    \end{minipage}%
    \\
    \rotatebox{90}{\hspace{-0.5cm}\small \textbf{RfD-Net}}&\hspace{-0.2cm}
    \begin{minipage}{\linewidth}
        \centering
    \includegraphics[width = 0.194\linewidth]{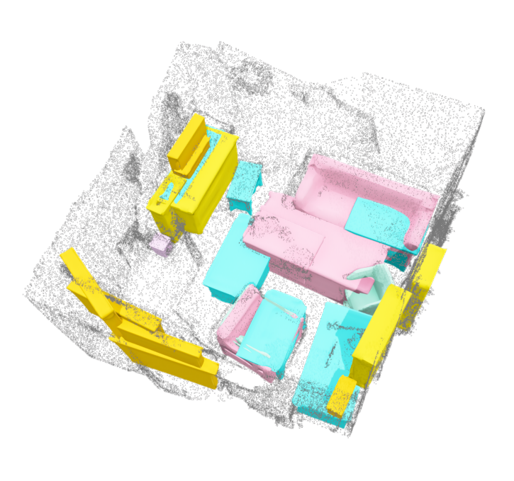}
    \includegraphics[width = 0.194\linewidth]{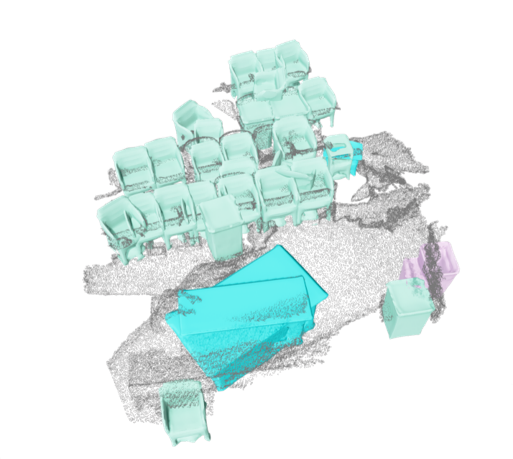}
    \includegraphics[width = 0.194\linewidth]{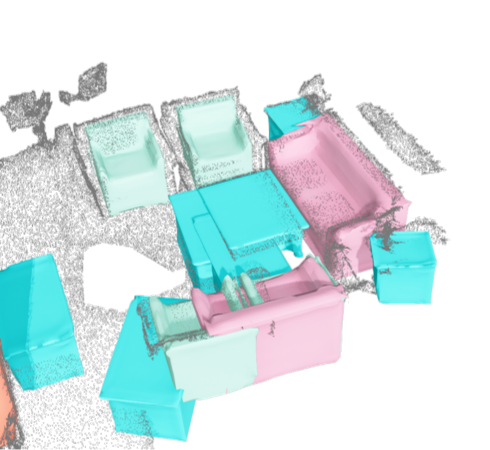}
    \includegraphics[width = 0.194\linewidth]{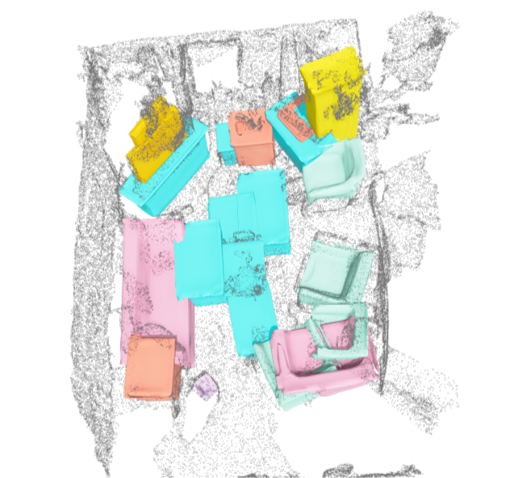} 
    \includegraphics[width = 0.194\linewidth]{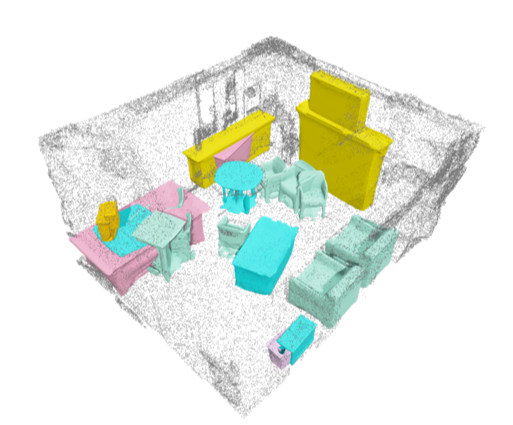}
     
    \end{minipage}%
    \\
    \rotatebox{90}{\hspace{-0.5cm}\small \textbf{DIMR}}&\hspace{-0.2cm}
    \begin{minipage}{\linewidth}
        \centering
    \includegraphics[width = 0.194\linewidth]{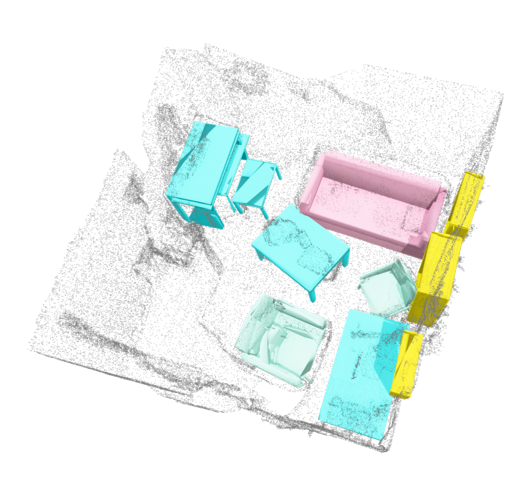}
    \includegraphics[width = 0.194\linewidth]{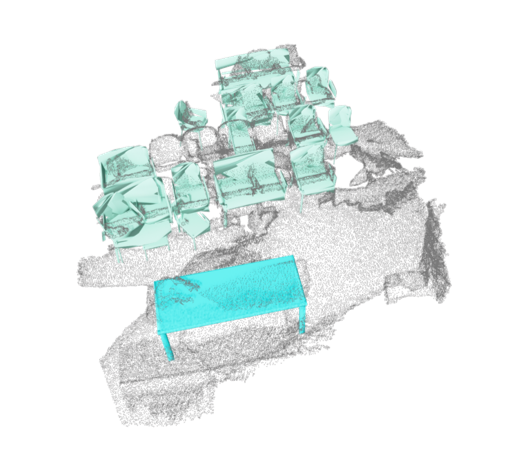}
    \includegraphics[width = 0.194\linewidth]{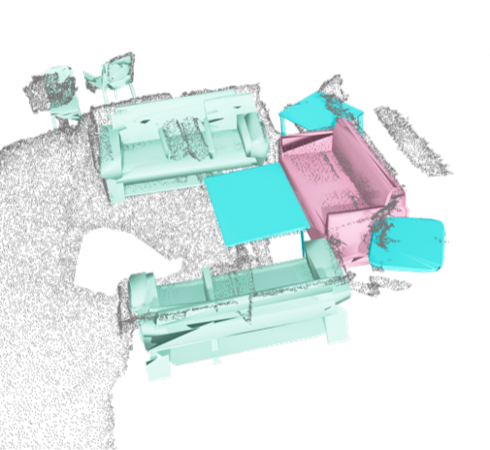}
    \includegraphics[width = 0.194\linewidth]{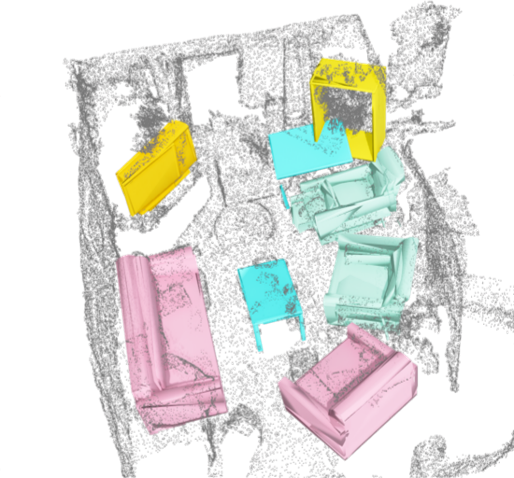}
    \includegraphics[width = 0.194\linewidth]{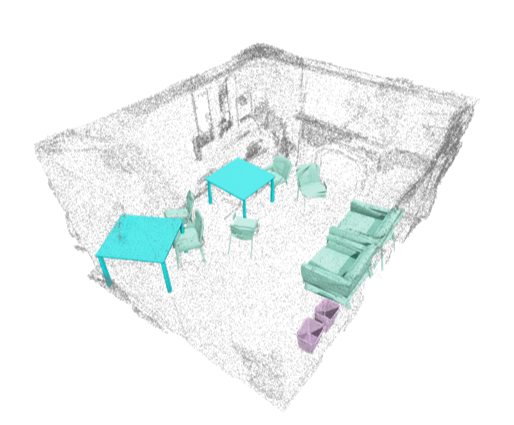}
    \end{minipage}%
    \\
    \rotatebox{90}{\hspace{-0.5cm}\small \textbf{Ours}}&\hspace{-0.2cm}
    \begin{minipage}{\linewidth}
        \centering
    \includegraphics[width = 0.194\linewidth]{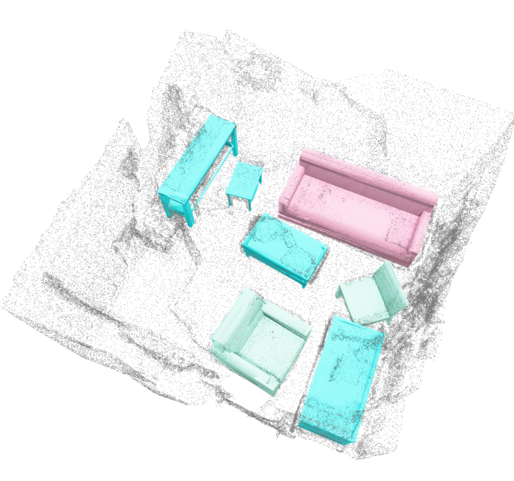}
    \includegraphics[width = 0.194\linewidth]{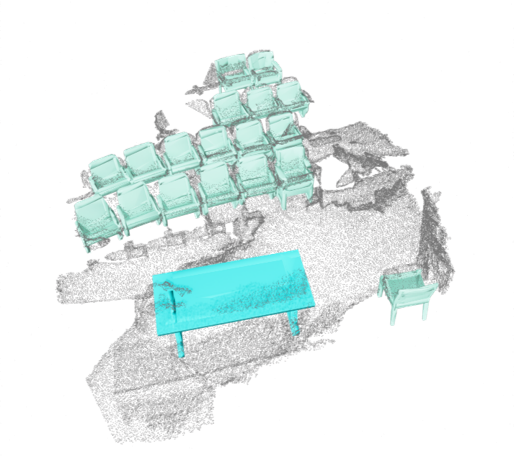}
    \includegraphics[width = 0.194\linewidth]{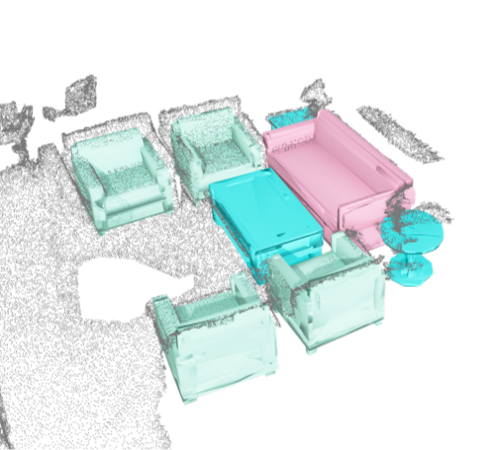}
    \includegraphics[width = 0.194\linewidth]{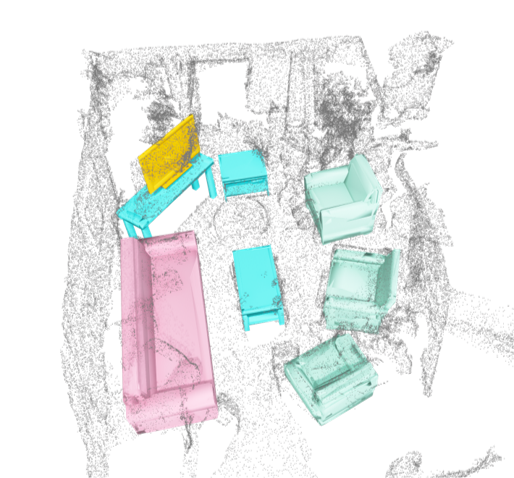}
    \includegraphics[width = 0.194\linewidth]{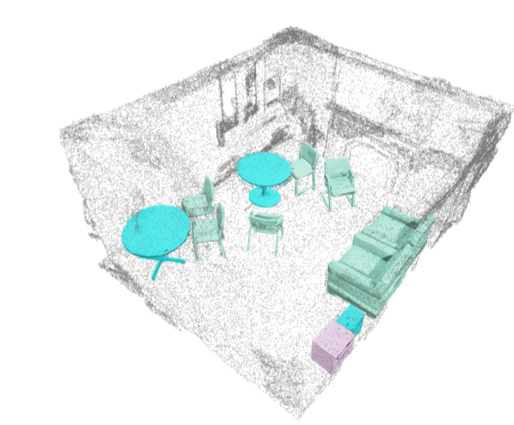}
    \end{minipage}%
    \\ \\
    % % \bottomrule
    % \multicolumn{2}{c}{\textbf{Semantic Scene Reconstruction}}
    % \\
    % % \bottomrule
    % \rotatebox{90}{\hspace{0.5cm}\small \textbf{Ours}}\rotatebox{90}{\hspace{0.3cm}\textbf{+Layout}}&\hspace{-0.2cm}
    % \begin{minipage}{\linewidth}
    % \centering
    % \includegraphics[width = \linewidth]{reconstruction_layout.png}
    % \end{minipage}%
    % \\
    % \bottomrule

\end{tabular}
\caption{Visualization comparison of scene semantic reconstruction on ScanNetv2.}
\label{fig:recon_comp} 
\end{figure*}

\begin{figure*}[h]
\setlength{\belowcaptionskip}{-1em}
    \centering
    \includegraphics[width = 0.9\linewidth]{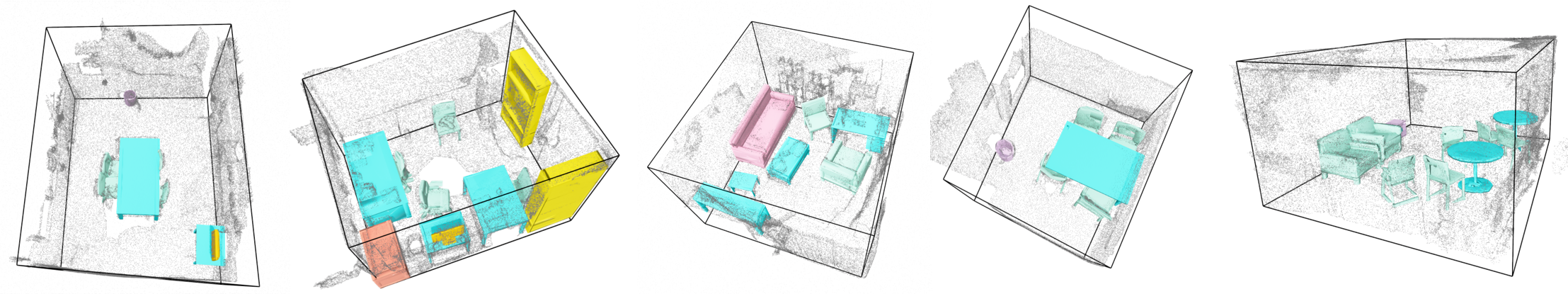}
    \caption{ Holistic scene understanding by our proposed AnchorRec. (The comparison methods didn't conduct layout estimation.)}
    \label{fig:layout}
\end{figure*}
% We illustrate the holistic scene understanding results in \cref{fig:layout}, including layout estimation, object detection, and instance reconstruction.

\textbf{Scene Reconstruction.}  We compare our method with the state-of-the-art works, \ie, RfD-Net \cite{nie2021rfd} and DIMR \cite{tang2022point}. RfD-Net predicts object shapes with the instance points segmented from proposals given by VoteNet, while DIMR infers object models from the instance points provided by a complex instance segmentation backbone. As shown in \cref{fig:recon_comp}, our AncRec can robustly recognize and reconstruct objects from noisy point clouds with accurate object localization and high-fidelity shape details. With the target-focused proposal feature learning offered by AncLearn in terms of object detection, our method generates fewer false positive models than the compared approaches. With more accurate detection results, our AncRec can leverage outlier-reduced instance priors sampled by AncLearn to infer more detailed object structures, \eg, the folding chairs are recovered in the second column of \cref{fig:recon_comp}. \cref{tab:rec} presents the quantitative comparison of different methods. Our AncRec outperforms other methods under all metrics. As the comparison methods didn't conduct layout estimation, we illustrate our holistic scene understanding results in \cref{fig:layout}. Considering that semantic scene reconstruction is a comprehensive problem, we further analyze the performance of our method on the key sub-tasks, \ie, scene parsing and object reconstruction, as follows.

\begin{table}[h]
\centering
\caption{Quantitative results on the ScanNetv2 dataset. We evaluate the reconstruction quality with mAP at different thresholds. (higher is better)}
\label{tab:rec}
\centering
\resizebox{\linewidth}{!}{
    \begin{tabular}{ccccccc}
        \toprule
        & IoU@0.25 & IoU@0.5 & CD@0.1  & CD@0.047 & LFD@5000 & LFD@2500\\ \midrule
        RfD-Net & 42.5 &  16.9 & 45.7  & 19.1 & 28.6 & 7.8\\
        DIMR    & 46.3 &  12.5 & 51.9 & 25.7 & 29.5 & 8.6\\
        AnchorRec & \bf{52.9} & \bf{18.9} & \bf{56.8}  & \bf{29.4} & \textbf{30.3} & \textbf{9.9}\\
        \bottomrule
\end{tabular}
}
\end{table}

\begin{figure*}[ht]
\centering
\begin{tabular}{c}
\hspace{1em}\includegraphics[width=0.9\linewidth]{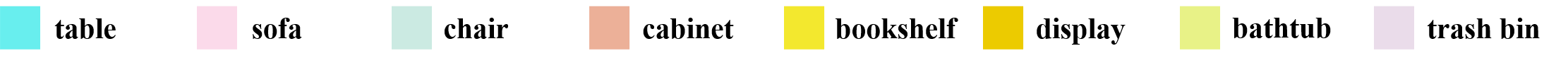}\\
    \includegraphics[width=0.9\linewidth]{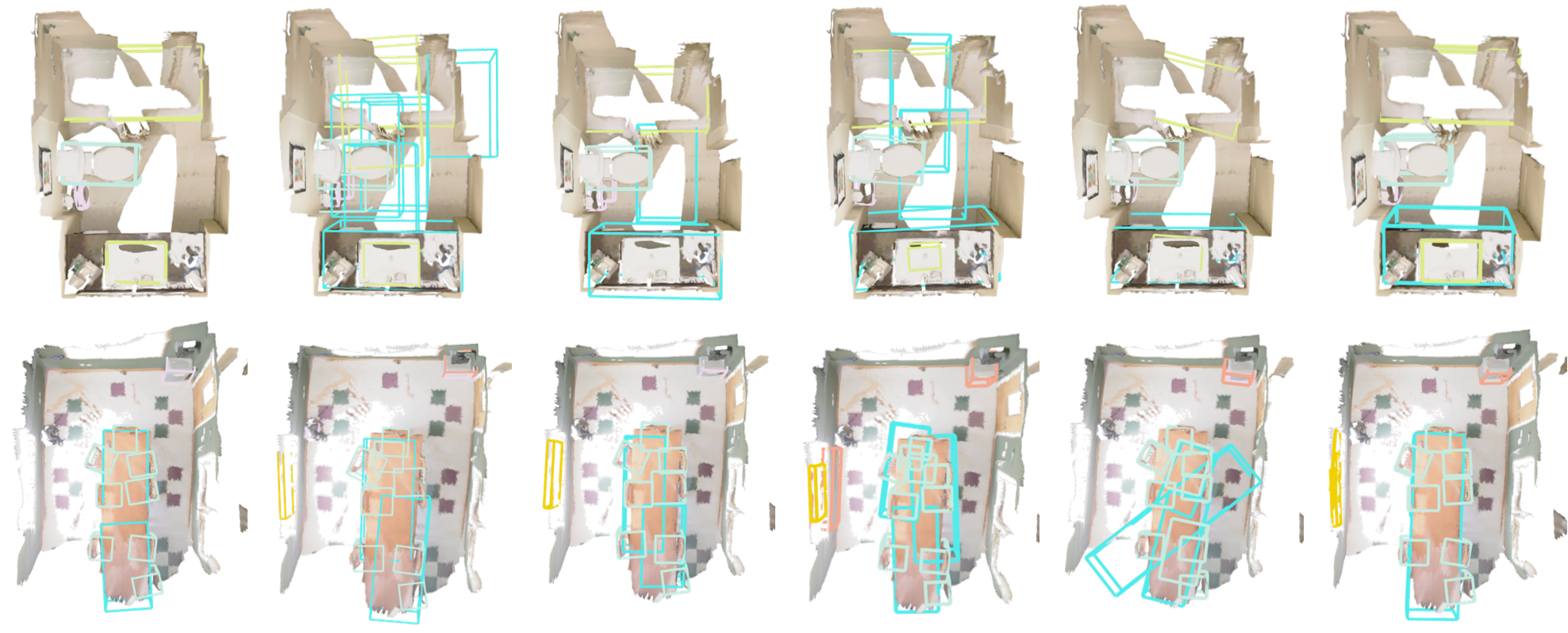}\\ 
    \includegraphics[width=0.9\linewidth]{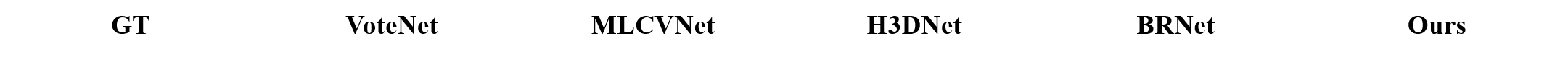}
\end{tabular}
\caption{Visualization comparison results of object detection on ScanNetv2}
\label{fig:object_detection}
\end{figure*}

\begin{figure*}[ht]
\centering
\begin{tabular}{cccc}
    \rotatebox{90}{~~~~~~~~~\textbf{GT}} &\includegraphics[height=6em]{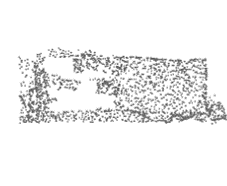} \includegraphics[height=6em]{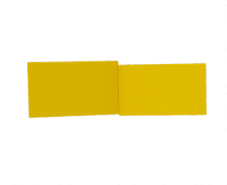}& \includegraphics[height=5.8em]{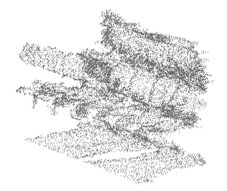} \includegraphics[height=5.8em]{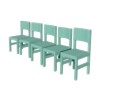}& \includegraphics[height=5.8em]{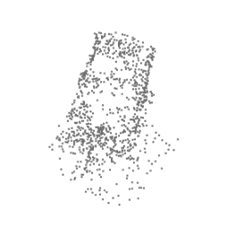} \includegraphics[height=5.8em]{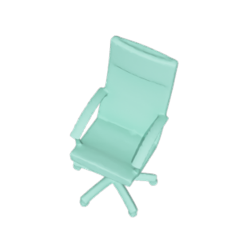}
    \vspace{-1em}\\
    \vspace{-1em}
    \rotatebox{90}{~~~~~~\textbf{RfD-Net}} &\includegraphics[height=5em]{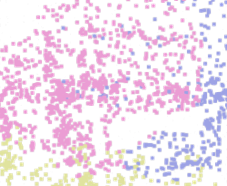} \includegraphics[height=5.8em]{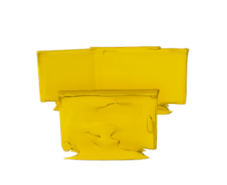}& \includegraphics[height=5.8em]{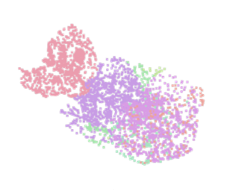} \includegraphics[height=5.8em]{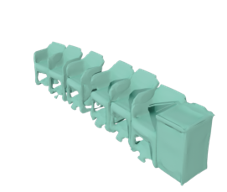}& \includegraphics[height=5.8em]{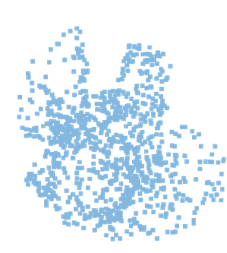} \includegraphics[height=5.8em]{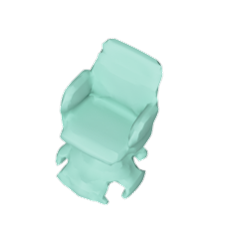}\\
    \vspace{-1em}
    \rotatebox{90}{~~~~~~\textbf{DIMR}} &\includegraphics[height=5.8em]{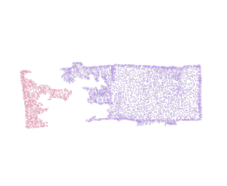} \includegraphics[height=5.8em]{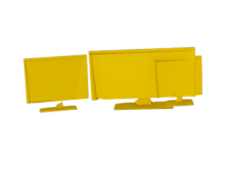}& \includegraphics[height=5.8em]{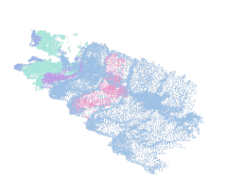} \includegraphics[height=5.8em]{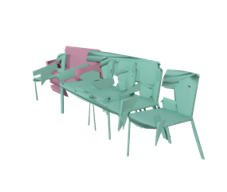}& \includegraphics[height=5.8em]{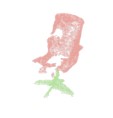} \includegraphics[height=5.8em]{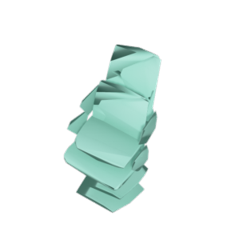}\\
    \vspace{-1em}
    \rotatebox{90}{~~~~~~~\textbf{Ours}} &\includegraphics[height=5.8em]{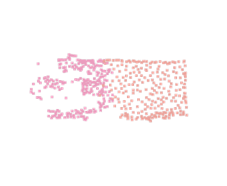} \includegraphics[height=5.8em]{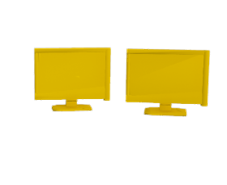}& \includegraphics[height=5.8em]{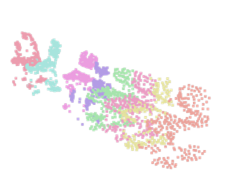} \includegraphics[height=5.8em]{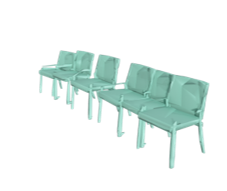}& \includegraphics[height=5.8em]{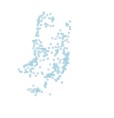} \includegraphics[height=5.8em]{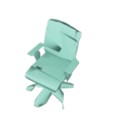}\\
\end{tabular}
\vspace{0.2em}
\caption{The influence of different instance point sampling strategies on object reconstruction. Points from the same instance are painted in the same color. Without the reliance on segmentation like RfD-Net and DIMR, our method can correctly separate adjacent objects and predict shape models.}
% \vspace{-1.5em}
\label{fig:sampling}
\end{figure*}

%numeric on OD
\begin{table*}[h]
\setlength{\belowcaptionskip}{-1em}
\centering
\caption{Quantitative results of object detection and layout estimation. The AP scores are measured with an IoU threshold of 0.5. Higher values mean better results.}
\begin{tabular}{lccccccccc|c}
    \toprule
    & \multicolumn{9}{c|}{\textbf{Object}} & \textbf{Layout}\\
    \midrule
    Method & table & sofa & chair & cabinet & bookshelf & display & bathtub & trash bin & mAP & F1-scores \\ \midrule
    VoteNet \cite{ding2019votenet} & 44.5 & 47.5 & 76.3 & 28.8 & 20.1 & 8.0 & 26.0 & 14.6 & 33.2 & -\\
    MLCVNet \cite{xie2020mlcvnet} & 50.5 & 46.6 & 81.8 & 25.5 & 19.2 & 7.8 & 25.3 & 19.2 & 34.5 & -\\
    H3DNet \cite{zhang2020h3dnet} & 53.6 & 52.4 & 76.0 & 25.6 & 27.2 & 9.4 & 20.7 & 10.0 & 34.4 & -\\
    BRNet \cite{cheng2021back} & 56.4 & 43.1 & 80.9 & 35.4 & 26.1 & 10.4 & 28.2 & 11.0 & 36.4 & -\\
    PQ-Transformer \cite{chen2022pq} & 52.7 & 52.5 & 81.1 & 32.1 & \bf{36.4} & 6.7 & 20.6 & 20.8 & 37.8 & 67.0 \\
    AnchorRec (Ours) & \bf{57.3} & \bf{56.1} & \bf{84.5} & \bf{36.9} & 29.8 & \bf{17.0} & \bf{34.0} & \bf{31.5} & \bf{43.4} & \bf{70.5}\\ 
    \bottomrule
\end{tabular}
\label{tab:od_results}
\end{table*}

%numeric on MESH
\begin{table*}[h]
\centering
\caption{Numeric results on object reconstruction. The AP scores are measured with CD at a threshold of 0.1. (Higher is better.)}
\begin{tabular}{lccccccccc}
    \toprule
    Method & table & chair & bookshelf & sofa & trash bin & cabinet & display & bathtub & mAP\\ \midrule
    RfD-Net \cite{nie2021rfd} & 25.51 & 82.11 & 32.53 & 44.21 & 44.74 & 28.37 & 65.51 & 42.57 & 45.70\\
    DIMR \cite{tang2022point} & 39.44 & 81.04 & \textbf{38.24} & 44.09 & \textbf{62.60} & 23.57 & \textbf{75.12} & 50.93 & 51.88\\
    AnchorRec (Ours) &\textbf{ 53.60} & \textbf{86.23} & 36.82 &\textbf{ 50.92} & 59.59 &\textbf{ 39.56} & 73.68 &\textbf{54.32} & \textbf{56.84}\\
    \bottomrule
    \vspace{-1em}
\end{tabular}
\label{tab:rec_per_class}
\end{table*}

% \vspace{-0.2em}
\textbf{Scene Parsing.} We compare AncRec with the state-of-the-art methods regarding 8-category object detection and layout estimation. The quantitative results in \cref{tab:od_results} show that our AnchorRec edges out the comparison methods by significant performance gains. Especially for the categories of the display,  bathtub, and trash bin, AncRec exceeds the second best by 6.6\%, 5.8\%, and 10.7\%  in precision measurement. The results indicate that, with adaptive grouping feature areas, AncLearn works effectively for detecting objects shaped irregularly, \eg, diverse-scaled bathtubs, thin displays, and tiny trash bins.  In \cref{tab:od_results}, we also compare our method with PQ-Transofrmer \cite{chen2022pq}, the current state-of-the-art scene parsing approach, in terms of layout estimation. The results demonstrate that our AncLearn is also applicable to localizing walls that are particularly large and thin. Visualized object detection result provided in \cref{fig:object_detection} further manifests that our AnchorRec can reliably delineate objects with compact bounding box prediction.

\textbf{Object Mesh Reconstruction.} 
In \cref{tab:rec_per_class}, we also evaluate the performance of different methods with respect to class-wise object reconstruction. The evaluation is based on how matchable the predicted object meshes are with the scene-aligned ground truths. The numeric results in \cref{tab:rec_per_class} show that AncRec achieves the best performance on 5 categories and the overall chamfer distance-based mAP score.

% \vspace{-0.5em}
\subsection{Ablation Studies}
% \vspace{-0.5em}

\textbf{AncLearn for Scene Parsing.} We study different settings of AncLearn for instance detection and report the results in \cref{tab: anchor-guided-grouper}. Compared to the baseline without AncLearn, the proposed AncLearn brings consistent improvement to object detection and layout estimation. Especially, when cooperating with the self-attention layer for layout estimation, AncLearn enables the dual-branch instance detector of AncRec to obtain the best performance in both parsing tasks.

\begin{table}[h!] 
	\centering
\vspace{-0.5em}	\caption{Ablation study of the proposed Anclearn. IoU@0.5 and F1 score  respectively reflect the performance of object detection and layout estimation. Higher values mean better results.}
	\label{tab:ablation}
	\centering
	\resizebox{\linewidth}{!}{
		\begin{tabular}{ccccc}
			\toprule
			OD anchor & layout anchor &layout SA & IoU@0.5 & F1 score   \\ \midrule
			&  &  & 41.12 & 60.62  \\ \midrule
			$\checkmark$  &  &  &  43.33(+2.21)  & 60.59(- 0.03)\\ 
			& $\checkmark$ &  &  40.21(-0.91)  & 62.61(+ 1.99)\\ 
			$\checkmark$  & $\checkmark$ & & 41.24(+0.12)  & 64.00(+3.38)\\
			& $\checkmark$ &  $\checkmark$  & 41.16(+0.04) & 65.34(+4.72) \\
			$\checkmark$  & $\checkmark$ & $\checkmark$ & \bf{43.38(+2.26)} & \bf{70.45(+9.83)} \\
			\bottomrule
	\end{tabular}}
	\label{tab: anchor-guided-grouper}
 \vspace{-1em}
\end{table}
\textbf{AncLearn for Object Reconstruction.} In the shape predictor of AncRec, AncLearn works to sample instance points as geometry priors for reconstruction. With the instance detector of AncRec, we compare AncLearn with the segmentation-based strategy used in RfD-Net and a box cropping sampling method. The quantitative comparison in \cref{tab:ablation_sampling_strategy} shows that AncLearn outperforms the other two approaches by providing instructive geometry priors with shape anchors in terms of all metrics.

\begin{table}[h] 
	\centering
	\caption{Comparison between different instance point sampling methods. Higher values mean better results.}
	\label{tab:ablation_sampling_strategy}
	\centering
	\resizebox{\linewidth}{!}{
		\begin{tabular}{cccc}
			\toprule
			& IoU@0.5 & IoU@0.25 & CD@0.1   \\ \midrule
			Segmentation \cite{nie2021rfd} &  17.4 & 51.4 & 55.8\\
			Box Cropping  &  17.7 & 51.9 & 55.5\\
			AncLearn (Ours)  &  \bf{18.9} & \bf{52.9} & \bf{56.8}\\
   \midrule
   \midrule
                & CD@0.047 & LFD@5000 &LFD@2500\\
                \midrule
                Segmentation \cite{nie2021rfd} &  29.1 & 29.1 & 7.8\\
			Box Cropping & 28.8 & 29.9 & 9.8\\
			AncLearn (Ours) & \bf{29.4} & \textbf{30.3} & \textbf{9.9}\\
			\bottomrule
	\end{tabular}}
  \vspace{-1em}
\end{table}

\subsection{Contributions of the Vote and Anchor Features}
\begin{figure}[h]
    \centering
    \includegraphics[width = \linewidth]{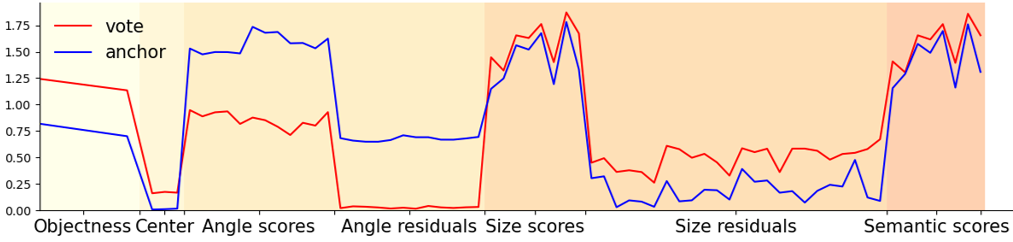}
    \caption{The learned weights for combining vote and anchor features.}
    \label{fig:weight}
\end{figure}

    As mentioned in \cref{sec:Anchor-guided Feature Grouper}, we perform instance detection by taking both vote and anchor features into account. To investigate the difference in their contributions to the detection parameters, we visualize the two weight vectors in \cref{eq:prediction_fusion} for perceptual comparison. As shown in \cref{fig:weight}, vote features have a larger impact on scoring objectness, indicating that the context encoded in vote features is useful in differentiating between object and non-object regions. Compared to vote features, anchor features contribute significantly more to angle prediction, which demonstrates the superiority of anchor-guided strategy in providing shape clues for learning shape-sensitive parameters. Considering the gap between the bounding box and the shape of objects, the vote features contribute more to predicting the bounding box center and size residuals.

\subsection{The Characteristics of the Learned Anchors} 
\begin{figure}[h]
\centering
    \resizebox{\linewidth}{!}{
    \begin{tabular}{cc}
        \centering
        \includegraphics[width = 0.16\linewidth]{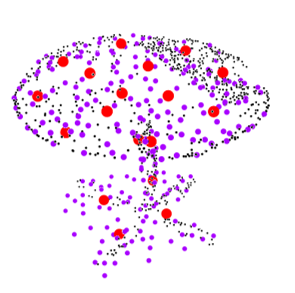}
        \includegraphics[width = 0.16\linewidth]{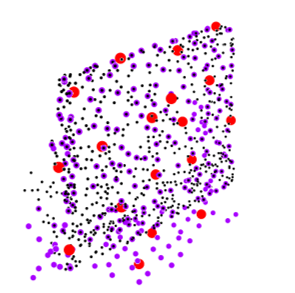}
        \includegraphics[width = 0.16\linewidth]{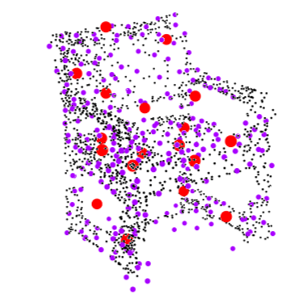}
        \includegraphics[width = 0.16\linewidth]{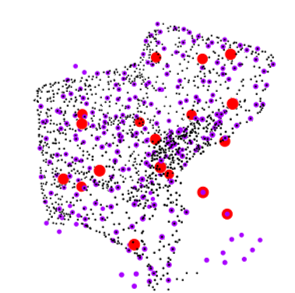}
        &
        \multicolumn{1}{|c}{\includegraphics[width = 0.16\linewidth]{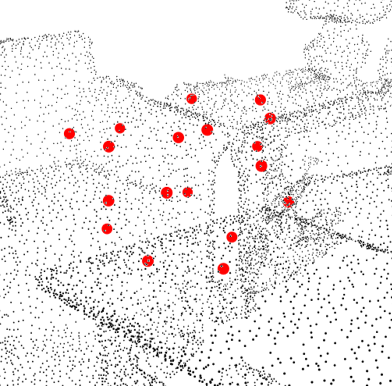}
        \includegraphics[width = 0.16\linewidth]{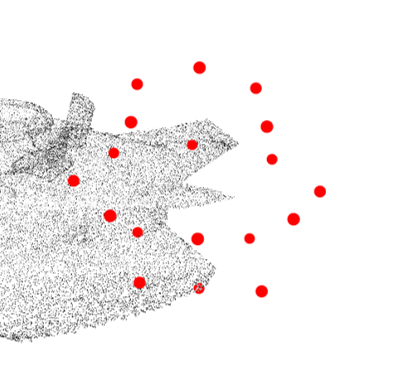}}\\
        \scriptsize Object Anchors & \scriptsize Non-object Anchors
        \end{tabular}
        }
\caption{The characteristics of the learned anchors. The anchors and sampled points are shown in red and purple respectively.}
\label{fig:anchor}
\end{figure}

The supervision with surface points derived from complete 3D models enables the anchors to depict object shapes, even with unobserved structures, which assists in object detection and reconstruction when some parts of objects are missing. There also exists a difference between anchor distribution characteristics for objects and non-object areas. As shown in \cref{fig:anchor}, the anchors and sampled points depict object shapes, while those for non-object areas scatter irregularly. This characteristic may play a role in differentiating objects from non-object regions.
% \begin{figure}[h]
%     \centering
%     \begin{tabular}{cc|cc|c}
%         \includegraphics[width=0.16\linewidth]{imgs/chair leg/1.png} & 
%         \includegraphics[width=0.16\linewidth]{imgs/chair leg/2.png} & 
%         \includegraphics[width=0.16\linewidth]{imgs/chair leg/4.png} & 
%         \includegraphics[width=0.16\linewidth]{imgs/chair leg/3.png} &
%         \includegraphics[width=0.16\linewidth]{imgs/chair leg/5.png}
%         \\    
%          \scriptsize instance points &  \scriptsize DIMR &  \scriptsize sampled  \scriptsize points & \scriptsize Ours & \scriptsize GT
%     \end{tabular}
%     \caption{Benefits of ground points around connection areas.}
%     \label{fig:ground points}
% \end{figure}

\section{Conclusions}
	In this paper, we introduce a shape anchor guided learning strategy (AncLearn) that is embedded into a reconstruction-from-detection learning system (AncRec) to handle the issue of noise interference in point-based holistic scene understanding. Extensive experiments demonstrate that AncRec achieves high-quality indoor semantic scene reconstruction. The quantitative and qualitative results show that AncRec outperforms current methods in terms of object detection, layout estimation, and shape modeling. The ablation studies convincingly verify that AncLearn can largely exclude noise from search space for reliable feature grouping and robust instance point sampling. In the future, it is promising to study the application of shape anchor guided learning strategy in other point-based 3D vision tasks.

\section*{Acknowledgment}
 This research was supported by the Fundamental Research Funds for the Central Universities of China under Grant 2042022dx0001, NSFC-projects under Grant 42071370, Wuhan University-Huawei Geoinformatics Innovation Laboratory, and the Open fund of Key Laboratory of Urban 
Land Resources Monitoring and Simulation, Ministry of Natural Resources
under Grant KF202106084.
	
	%%%%%%%%% REFERENCES
	{\small
		\bibliographystyle{ieee_fullname}
		\bibliography{final}
	}

\end{document}